\documentclass{article}

\usepackage{microtype}
\usepackage{graphicx}
\usepackage{subfigure}
\usepackage{booktabs} 
\usepackage{hyperref}
\usepackage{amsmath}
\usepackage{amssymb}
\usepackage{mathtools}
\usepackage{amsthm} 
\usepackage[accepted]{icml2023}
\usepackage[capitalize,noabbrev]{cleveref}

\theoremstyle{plain}
\newtheorem{theorem}{Theorem}[section]

\theoremstyle{definition}

\theoremstyle{remark}

\icmltitlerunning{Information Content based Exploration}
\usepackage{icml2023}
\begin{document}

\twocolumn[
\icmltitle{Information Content based Exploration}
\icmlsetsymbol{equal}{*}

\begin{icmlauthorlist}
\icmlauthor{Jacob Chmura}{equal,comp}
\icmlauthor{Hasham Burhani}{equal,comp}
\icmlauthor{Xiao Qi Shi}{equal,comp}
\end{icmlauthorlist}

\icmlaffiliation{comp}{RBC Capital Market, Toronto, Canada}
\icmlcorrespondingauthor{Jacob Chmura}{jacob.chmura@rbccm.com}
\icmlcorrespondingauthor{Hasham Burhani}{hasham.burhani.rbccm.com}
\icmlcorrespondingauthor{Xiao Qi Shi}{xiaoqi.shi@rbccm.com}

\icmlkeywords{Reinforcement Learning, Exploration, Information, Entropy}

\vskip 0.3in
]

\newcommand\blfootnote[1]{%
  \begingroup
  \renewcommand\thefootnote{}\footnote{#1}%
  \addtocounter{footnote}{-1}%
  \endgroup
}
\blfootnote{RBC Capital Market, Toronto, Canada. Correspondence to: Jacob Chmura (jacob.chmura@rbccm.com), Hasham
Burhani (hasham.burhani.rbccm.com), Xiao Qi Shi (xiaoqi.shi@rbccm.com)}

\begin{abstract}
Sparse reward environments are known to be challenging for reinforcement learning agents. In such environments, efficient and scalable exploration is crucial. Exploration is a means by which an agent gains information about the environment; we expand on this topic and propose a new intrinsic reward that systemically quantifies exploratory behaviour and promotes state coverage by maximizing the information content of a trajectory taken by an agent. We compare our method to alternative exploration-based intrinsic reward techniques, namely Curiosity Driven Learning (CDL) and Random Network Distillation (RND). We show that our information-theoretic reward induces efficient exploration and outperforms in various games, including Montezuma’s Revenge – a known difficult task for reinforcement learning. Finally, we propose an extension that maximizes information content in a discretely compressed latent space which boosts sample-efficiency and generalizes to continuous state spaces.
\end{abstract}

\section{Introduction}

\noindent Given an agent without knowledge of the environment dynamics, a
fundamental question is: what should the agent learn
first? When the task is also unknown, the question becomes increasingly complex \cite{moerland2020model}\cite{riedmiller2018learning}. In reward-dense environments, the agent receives a continuous gradient signal that guides learning through interaction \cite{mnih2013playing}\cite{achiam2017surprise}\cite{hong2018diversity}\cite{haarnoja2018soft}. However, for many problems of interest, the most faithful reward specification is sparse, degrading the sample efficiency with which online algorithms can learn desirable policies \cite{openaimontezuma}\cite{vecerik2017leveraging}\cite{singh2019end}\cite{sekar2020planning}. \\
\begin{figure}[ht]
    \centering
    \includegraphics[width=0.45 \textwidth]{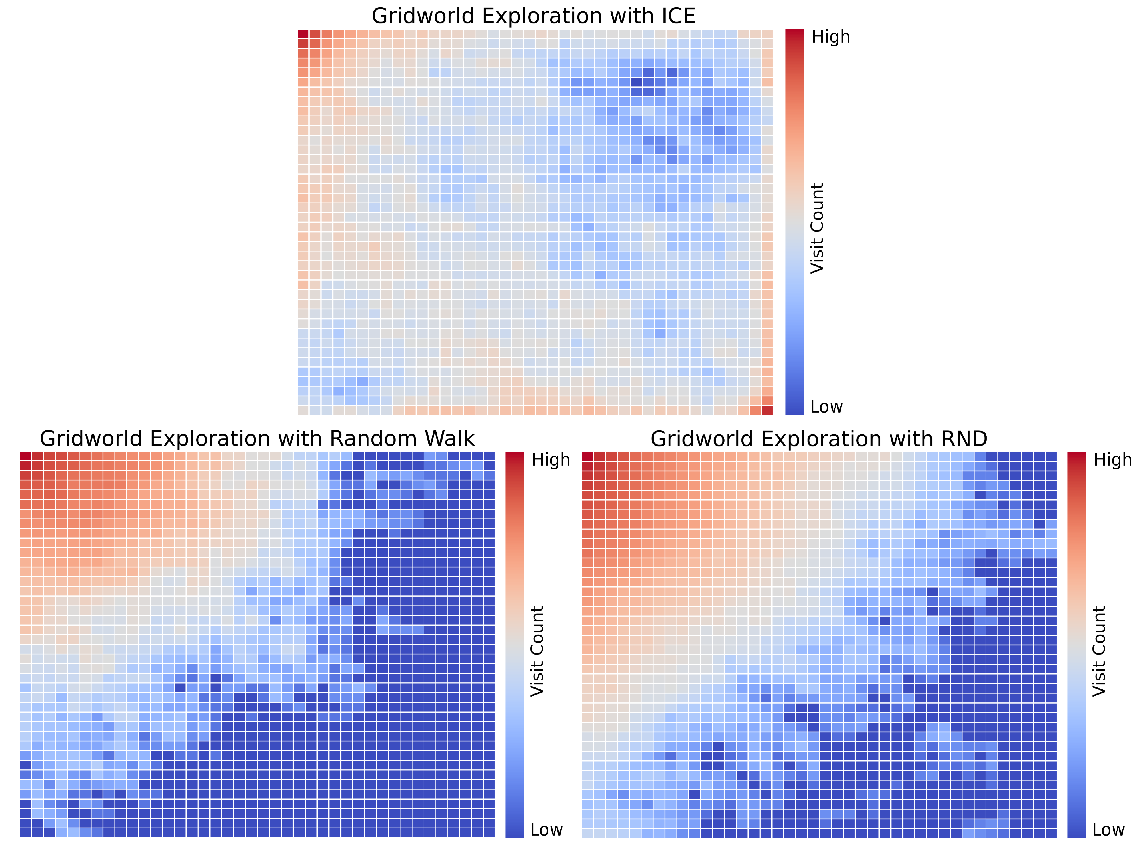}
    \caption{ICE (top) efficiently explores the no-reward grid environment. State visitation is more uniform as compared to Random Walk (bottom left) and RND (bottom right), which have low visitation density in the area opposite the agent start location (top left of grid). ICE incentives trajectories with distinct states, contrary to action space entropy which often revisiting the same states multiple times in a cyclic fashion.}
    \label{fig:gridcount}
\end{figure} 

\begin{figure}[ht]
    \centering
    \includegraphics[width=0.45\textwidth]{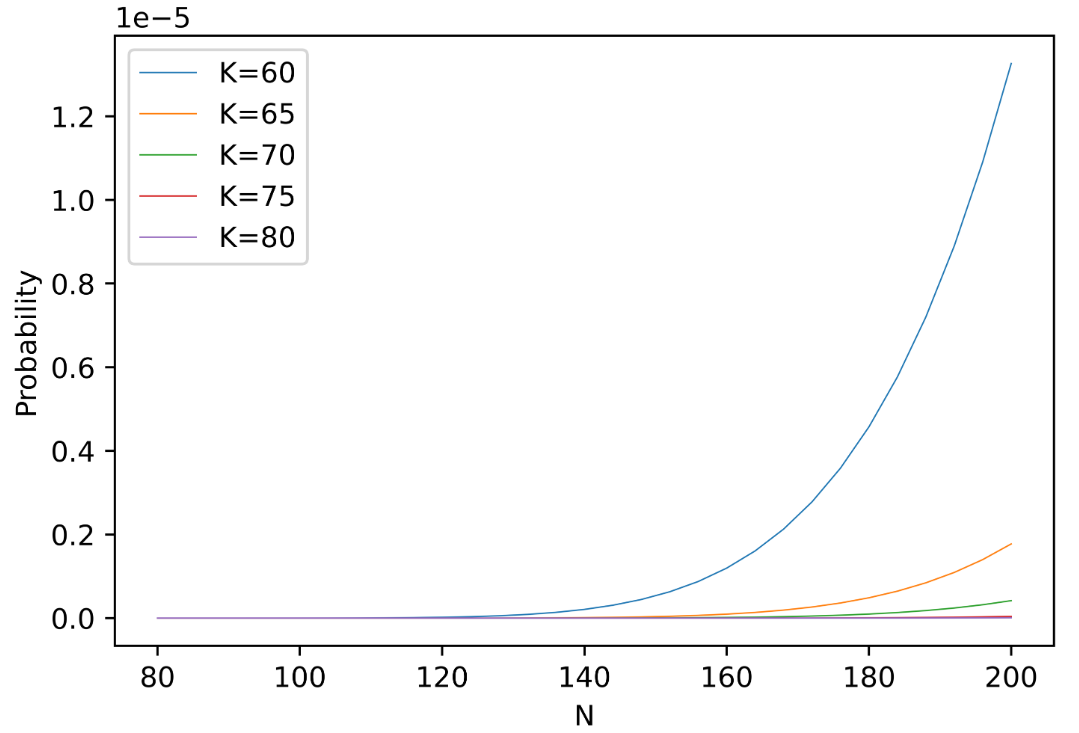}
    \caption{A random exploration strategy works with fairly high probability if the solution can be achieved in a small distance traveled from the origin. As can be seen above, as the distance from origin \(K\) increases, the probability of success decreases exponentially for the same episode length \(N\). See \ref{ss:a2} in the Appendix for derivation.}
    \label{fig:randomprob}
\end{figure}

\noindent To date, there have been a variety of ways in which this has been addressed. A common approach is to leverage domain knowledge to artificially shape the reward function into one amenable to credit assignment \cite{reward_shape}. However, engineered rewards require extensive domain knowledge, suffer from human cognitive bias, and often lead to unintended consequences \cite{faulty}. Another popular method has been to induce random exploration through entropy maximization in the action space \cite{ahmed2019understanding}. This heuristic induces random walk behavior, as seen in Fig.\ref{fig:gridcount}, and therefore suffers from scalability issues in high-dimensional state spaces: the probability of traveling far from the starting state is reduced exponentially (Fig.\ref{fig:randomprob}). 

\noindent At the same time, curiosity-driven learning enables humans to explore complex environments efficiently and integrate novel experiences in preparation for future tasks \cite{human_curiosity}. Recent work has focused on emulating curiosity in reinforcement learning agents through intrinsic reward bonuses such as pseudo-counts \cite{psuedocounts} and variational information gain \cite{vime}. CDL uses a form of state dynamics prediction error to entice the agent to visit unfamiliar states \cite{CFD}. Similarly, RND uses random features as a learning target for a curiosity model, the error of which is used as an intrinsic reward; neural network prediction error is observed to be a good proxy for novelty and hence quantifies exploration progress \cite{RND}. We contribute to the family of curiosity-inspired exploration methods by proposing an information theory-driven intrinsic reward that induces effective exploration policies without introducing auxiliary models or relying on approximations of environment dynamics.

\noindent Information Content-based Exploration (ICE) introduces an intrinsic reward that maximizes information gain in the state space. In each episode of interaction with the environment, we aggregate trajectory statistics to approximate the entropy of the state visitation distribution and reward the agent for relative improvements to this measure. This approach formalizes the exploration process as seeking low-density states but can easily accommodate prior knowledge of the environment by replacing entropy with KL-divergence to a desired state density, similar to \cite{ssm}. 

\noindent We run several experiments comparing ICE’s performance against RND and CDL using A3C \cite{mnih2016asynchronous} as our base RL algorithm to show that ICE significantly outperforms RND and CDL in various environments. We also observe that ICE exploration exhibits trajectories akin to depth-first search. In most environments, this offers the best opportunity to find extrinsic rewards, where the agent must decisively commit to a path through the environment rather than dithering locally via action entropy. Finally, we discuss extensions to our approach that generalize to continuous state spaces by maximizing information content on a hashed latent space extracted from an auto-encoder architecture.

\section{Background}

\noindent We briefly cover A3C in section \ref{ss:A3C}, and introduce the Source Coding Theorem which provides the theoretical justification for using entropy in the state space for exploration in ICE. In \ref{ss:CDL} and \ref{ss:RND}, we cover Curiosity Driven Learning and Exploration by Random Network Distillation algorithms. 

\subsection{Actor Critic Algorithm} \label{ss:A3C}
We consider a parametric policy $\pi_{\theta} \in \Pi := \{\pi_{\theta}: \theta \in \Theta\}$ interacting with a discounted Markov Decision Process $\mathcal{M} = $(\(\mathcal{S}, \mathcal{A}, \gamma, \mathcal{P}, r\)). At each time step \(t\), the policy outputs an action $a_t \sim \pi_{\theta}(\cdot | s_t)$ that leads to the next state $s_{t+1}$ according to the transition kernel $\mathcal{P}(s_{t+1}|s_t, a_t)$. The environment also provides an extrinsic reward $r_t \sim (s_t, a_t)$. The objective is to maximize the expected discounted cumulative reward \(\mathcal{J}(\pi)=\mathbb{E}_{\pi}(\sum_{t=0}^{\mathcal{T}} \gamma^{t}r_{t})\). The critic is an auxiliary model with parameters $\phi \in \Phi$ that reduces the variance of the policy gradient by approximating the value function: $\hat{V}_{\phi}^{\pi}(s_t) \approx \mathbb{E}_{\pi}[\sum_k\gamma^kr_{t+k+1}|s_t]$.

The Actor-Critic method maximizes the objective \(\mathcal{J}\) by optimizing the model’s weights based on value loss, policy loss, and entropy loss, which have the form
\begin{equation} \label{eq:A3C}
\begin{split}
L_{value, t} &= \alpha_{value}(r_{t} + \gamma V_{\phi}(s_{t+1}) - V_{\phi}(s_{t}))^{2} \\
L_{policy, t} &= \alpha_{policy}(-A_{t}\log \pi_{\theta}(a_t | s_{t})) \\
L_{entropy, t} &= \alpha_{entropy}(-\mathbb{H}(\pi_{\theta}(\cdot | s_{t})))
\end{split}
\end{equation}
\noindent where \(\alpha_{value}, \alpha_{policy}, \alpha_{entropy}\) are the weight coefficients of the three losses,  \(\pi_{\theta}(a_t|s_{t})\) is the probability of the selected action policy, \(\mathbb{H}(\pi_{\theta}(\cdot | s_{t}))\) is the entropy of the policy. \\
\noindent In addition, \(k\)-step Advantage estimate \(A_{t}\) is defined to be
\begin{equation} \label{eq:Advantage}
A_{t} = \sum_{i=0}^{k-1}(\gamma^{i}r_{t+i}) + \gamma^{k}V_{\phi}(s_{t+k}) - V_{\phi}(s_{t})
\end{equation}
\noindent In the case where the intrinsic reward is present, the reward \(r_{t}\) is the weighted sum of intrinsic and extrinsic reward,
\begin{equation} \label{eq:Reward}
r_{t} = r^{extrinsic}_{t} + \beta r^{intrinsic}_{t}
\end{equation}
where \(\beta\) is the weight coefficient of the intrinsic reward.

The k-step state distribution induced by a policy $\pi$ is given by:
\begin{align}
    d_{k, \pi}(s) &= \mathbb{E}_{\substack{s_1 \sim \rho_0(\cdot) \\ a_t \sim \pi(\cdot|s_t) \\ s_{t+1}\sim \mathcal{P}(\cdot|s_t, a_t)}}[\frac{1}{k}\sum_{t=1}^k 1(s_t=s)]
\end{align}

\subsection{Source Coding Theorem} \label{ss:SCT}
Given a list of discrete elements \(L = \{l_{0}, l_{1}, ..., l_{t}\}\), \(l_{i}\in\mathbb{R}^{1}\), the Source Coding Theorem states that \(L\) can be compressed into no less than \(H^{L}_{t} \times (t+1)\) bits, where \(H^{L}_{t}\) is the entropy of \(L\) up to step \(t\).
\begin{theorem} [Source Coding Theorem \cite{shannon1948mathematical}, \(D=1\)]
A single random variable (D=1) with entropy \(H\) can be compressed into at least \(H\) bits without risk of information loss.
\end{theorem}

We note that the joint entropy is sub-additive, with equality when the random variables are independent \cite{natural}:
\begin{equation}
    H(X_1, ..., X_n) \leq \sum_i H(X_i)
\end{equation}

We also note that up to a constant, the entropy of a discrete random variable $X: \mathcal{X} \to \{1, ..., K\}$ distributed according to $p$ is given by $-D_{KL}(p, \mu)$ where $\mu$ is the uniform distribution on $\mathcal{X}$, and $D_{KL}$ is the KL divergence:
\begin{align}
    D_{KL}(p, \mu) &= \sum_{k=1}^Kp(X=k)log\frac{p(X=k)}{K^{-1}} \\
    &= log K + \sum_{k=1}^Kp(X=k)logp(X=k) \\
    &= log K - \mathbb{H}(X)
\end{align}

Hence, maximizing the state distribution entropy is equivalent to minimizing the reverse KL to the uniform distribution over state.

\subsection{Curiosity Driven Learning (CDL)} \label{ss:CDL}
This branch of exploration provides an intrinsic reward based on the agent's familiarity with the current state; curiosity-driven Exploration by Self-supervised Prediction \cite{pathak2017curiosity} is one of the popular approaches. It formulates the intrinsic reward by using three neural networks to estimate state encoding, forward dynamics, and inverse dynamics. The first neural network \(f_{encode}(s)\)  encodes \(s_{t}\mapsto g_{t}\) and \(s_{t+1}\mapsto g_{t+1}\). Then a second neural network \(f_{forward}(g_{t},a_{t})\) estimates the next state encoding \(\hat{g}_{t+1}\). Lastly, the third neural network \(f_{inverse}(g_{t},g_{t+1})\) estimates the transition action distribution \(\hat{a_{t}}\). The training loss and intrinsic reward have the form
\begin{equation} \label{eq:CDL}
\begin{split}
L_{forward_{t}} &= \alpha_{forward}MSE(g_{t+1}-\hat{g_{t+1}}) \\
L_{inverse_{t}} &= \alpha_{inverse}CrossEntropy(\hat{a_{t}},a_{t}) \\
r_{intrinsic_{t}} &= \beta L_{inverse_{t}}
\end{split}
\end{equation}
The agent will naturally favor states that it cannot accurately predict the transition action from \(s_{t}\) to \(s_{t+1}\). 
\subsection{Exploration by Random Network Distillation (RND)} \label{ss:RND}
Exploration by Random Network Distillation \cite{RND} follows the same design philosophy as Curiosity Driven Learning. It formulates the intrinsic reward by using two neural networks. The first neural network \(f_{random}(s)\), encodes \(s_{t}\mapsto g_{t}\). It is randomly initialized and will not get updated. The second neural network \(f_{encode}(s)\), encodes \(s_{t}\mapsto \hat{g}_{t}\). The training loss and intrinsic reward have the form
\begin{equation} \label{eq:RND}
\begin{split}
& L_{encode_{t}} = \alpha_{encode}MSE(g_{t+1},\hat{g_{t+1}}) \\
& r_{intrinsic_{t}} = \beta L_{inverse_{t}}
\end{split}
\end{equation}
where \(\alpha_{encode}\) and \(\beta\) are coefficients. This method encourages the agent to explore in states where \(f_{encode}\) cannot accurately predict the random encoding.

\section{Information Content-based Exploration (ICE)}

\subsection{Motivation}

While interacting with the environment, we view the t-step state sequence $s_1, ..., s_t$ as a 1-sample Monte-Carlo approximation to the t-step state distribution $\hat{d}_{t, \pi}(s) \approx d_{t, \pi}(s)$ induced by the agent $\pi$. Supposing that our state space $\mathcal{S}$ is isomorphic to $\textbf{K}^D := \{1, 2, .., K\}^D$, then we can compute \footnote{An efficient vectorized implementation can be found in the supplemental code.} the empirical entropy of the t-step state sequence $\hat{d}_{t, \pi}(s)$:
\begin{equation}
    \mathbb{H}(\hat{d}_{t, \pi}(s)) = -\sum_{x \in \textbf{K}^D}\hat{p}(x) log \hat{p}(x)
\end{equation}

in $\mathcal{O}(t \cdot K^{D})$. For computational reasons, we view each state trajectory as composed of \(D\) \emph{independent} state elements (e.g., frames of a video made up of \(D\) pixels) which reduces the complexity to $\mathcal{O}(t \cdot K \cdot D)$:
\begin{equation}
    \sum_{x \in \textbf{K}^D}\hat{p}(x) log \hat{p}(x) \approx \sum_{d \in D}\sum_{k \in \textbf{K}} \hat{p}(x_d=k) log \hat{p}(x_d=k) 
\end{equation}

We provide an intrinsic reward to the agent based on the \emph{relative improvement} of this measure across time steps. Thus, we optimize for the approximate derivative of state entropy. We view several benefits to using the relative improvement measure:
\begin{itemize}
    \item Intrinsic stochasticity (e.g. persistent white noise) has high Shannon entropy but low fundamental information content (since the data is non-compressible). The relative improvement measure nullifies such entropy sources.
    \item Let $\delta_t \geq 0$ be the sub-additivity gap introduced by assuming our state-sequence $s_1, ..., s_t$ is independent. If $\delta_t$ is relatively constant across time then the relative improvement measure cancels out the overestimation of state entropy introduced by our independence assumption.
\end{itemize}

In \ref{ss:algo} we give an efficient algorithm for computing the intrinsic ICE reward, then, in \ref{ss:discretization} we discuss a principled approach to state discretization, which enables us to compute ICE in continuous state spaces.

\subsection{Algorithm}  \label{ss:algo}
Let us define a \emph{trajectory}, \(X_{t}\), to be the list of states collected from the same episode up to step \(t\), \(X_{t} = [s_{0},s_{1},..., s_{t}]\), where each state \(s_{t}\) is a collection of \emph{state elements},  \(s_{t}=[s_{t}^{0},s_{t}^{1},..., s_{t}^{D-1}]\in\mathbb{R}^{D}\). To collect the total information in a trajectory, we first calculate the information content received over each \emph{state element} in a trajectory. \\

\begin{figure}[ht]
    \centering
    \includegraphics[width=0.45\textwidth]{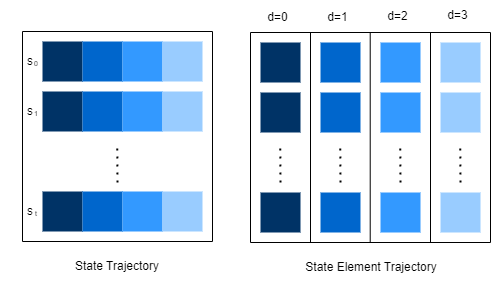}
    \caption{A trajectory is the collection of all states traversed by the agent in sequence. State element trajectory is the collection of elements from the same position in all states in a trajectory.}
    \label{fig:statetrajectory}
\end{figure} 

\noindent Consider an arbitrary state element \(d\) in a trajectory up to step \(t\), \(X^{d}_{t} = [s_{0}^{d},s_{1}^{d},..., s_{t}^{d}]\). We can calculate the count for each unique value within \(X^{d}_{t}\).
\begin{equation} \label{eq:unique_count}
q^{d}_{t} = CountUnique(X^{d}_{t})
\end{equation}

\noindent Next, we can calculate the probability of each unique value occurrence within \(X^{d}_{t}\).
\begin{equation} \label{eq:prob_occur}
p^{d}_{t} = \frac{q^{d}_{t}}{t+1}
\end{equation}

\noindent For example, if \(X^{d}_{t=4}=[a,a,b,a,c]\), then \(q^{d}_{t=4}=[3,1,1]\) and \(p^{d}_{t=4}=[3/5,1/5,1/5]\). \\

\noindent We can then obtain the entropy for \(X^{d}_{t}\) by applying Shannon's entropy to \(p^{d}_{t}\).
\begin{equation} \label{eq:unit_entropy}
H^{d}_{t} = Entropy(p^{d}_{t}) =  -\sum_{k} p^{d}_{t}[k]\log_{2} p^{d}_{t}[k]
\end{equation}
\(H^{d}_{t}\) measures the information content of the \(d^{th}\) state element trajectory. The information content of the full trajectory can be calculated by summing over all state elements.
\begin{equation} \label{eq:traj_entropy}
H_{t} = \sum_{d} H^{d}_{t}
\end{equation}
We then denote \(r^{intrinsic}_{t}\) as
\begin{equation}
r^{intrinsic}_{t} = H_{t} - H_{t-1}
\end{equation}

Intuitively, \(r^{intrinsic}_{t}\) represents the amount of additional information \(s_{t}\) brings to the trajectory \([s_{0}, s_{1}, ..., s_{t-1}]\). Note that \(\sum r^{intrinsic}_{t} = H_{T}\), which is the total information content of the entire trajectory. A numerical example of \(r^{intrinsic}_{t}\) calculation can be found in Appendix \ref{ss:example}. \\

\begin{algorithm}[tb]
   \caption{Information Content-based Intrinsic Reward}
   \label{alg:full}
\begin{algorithmic}
   \STATE {\bfseries Input:} Model \(M\)
   \REPEAT
       \STATE {\bfseries Input:} state \(s_{0}\), {$H_{last} \gets 0$}, {$t \gets 0, \ done \gets False$}
       \STATE {\bfseries Input:} unique count dictionary \(q\)
       \STATE {$q \gets Update(q, s_{0})$}, from Eq.\ref{eq:unique_count}
       \STATE {$H_{last} \gets CalculateICE(q)$},  from Eq.\ref{eq:traj_entropy}
       \WHILE {not \(done\)}
            \STATE {$a_{t} \gets M(s_{t})$}
            \STATE {$s_{t+1},r^{extrinsic}_{t},done \gets env.step(a_{t})$}
            \STATE {$q \gets Update(q, s_{t+1})$}
            \STATE {$H_{current} \gets CalculateICE(q)$}
            \STATE {$r^{intrinsic}_{t} \gets H_{current}-H_{last}$}
            \STATE {$r_{t} \gets r^{extrinsic}_{t} + \beta r^{intrinsic}_{t}$}
            \STATE {$H_{last} \gets H_{current}$}
            \STATE {$t \gets t+1$}
       \ENDWHILE
       \STATE {Update model \(M\)}
   \UNTIL{Model \(M\) Converge}
\end{algorithmic}
\end{algorithm}

\noindent This formulation of the \(r^{intrinsic}\) requires no estimation, as it can be calculated directly from the trajectory of states. By maintaining count arrays of size $D \times K$, we can efficiently implement Alg. $\ref{alg:full}$ using vectorized CPU or GPU operations. See the supplementary Material for our implementation.

\subsection{State Exploration vs Action Exploration} \label{ss:discussion}

\noindent ICE is a state space exploration algorithm and assumes the environment to be an observable Markov Decision Process. This assumption is necessary because ICE measures the information content of the observable state trajectory. \\

\noindent ICE encourages the agent to efficiently explore the trajectory containing the most information content at the cost of disincentivizing the agent from pursuing trajectories with low information. Therefore, ICE offers no guarantee that every state in the decision process will be visited. On the other hand, action space exploration (Eq.\ref{eq:A3C}) encourages the agent to take actions based on uniform distribution. Albeit inefficient, it does guarantee that the agent will eventually visit all states in an environment. \\

\begin{figure}[ht]
    \includegraphics[width=0.45\textwidth]{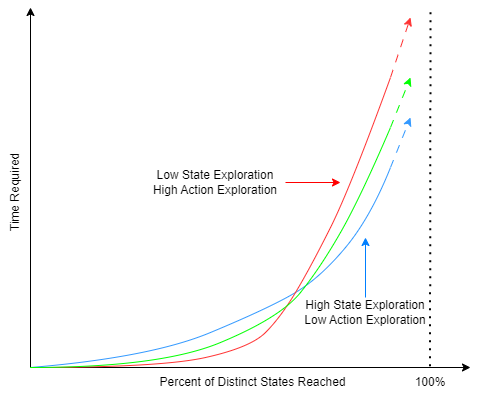}
    \caption{As we increase the action exploration and decrease the state exploration, it takes significantly longer to reach the same amount of distinct states (blue line). In the case where we decide to only apply state exploration, the agent will be able to efficiently follow the trajectory with very rich information content, at the cost of never visiting states in the low information content trajectories. As a result, such an agent will almost never reach all the states (red line).} 
    \label{fig:Exploration}
\end{figure}

\noindent State space exploration is fundamentally different than action space exploration. They are complements to each other. In the case where the agent is driven only by the action of space exploration, the agent can reach all reachable states, but the expected amount of time required can be very long. On the other hand, an agent driven only by state space exploration can reach more distinct states in significantly less amount of time, but it is no longer guaranteed to reach all reachable states. The probability of reaching a reward is proportional to the percent of all possible states reached. As shown in Fig.\ref{fig:Exploration}, The agent with high state exploration and low action exploration can potentially reach rewards in a shorter amount of time but with a possibility that it may never reach all the rewards. Therefore, it is important to balance the weighting of these two types of explorations in order to explore optimally, illustrated in (Fig.\ref{fig:Exploration}) and \ref{ss:gridworldwall}.

\subsection{ICE in a Learned Latent Space} \label{ss:discretization}

A trajectory's information content depends on how we choose to discretize the state space; naturally, different discretization methods will change the absolute information content. Selecting the appropriate discretization is only possible with prior knowledge of the environment. We propose to address this by using an auto-encoder to learn a compressed, low-dimensional discrete representation upon which information is maximized. Computing ICE on the latent manifold decouples the ICE formulation from the dimensionality of the observation space. The information bottleneck principle \cite{ib} allows us to trade between \emph{sufficiency} and \emph{minimality} in our discrete representation. 

Since the KL divergence is parameterization-invariant, we can be sure that insofar as our auto-encoder learns an invertible representation of the state space, computing ICE in latent space should lead to desirable exploration behaviour as in discrete settings. We validate this procedure in \ref{test:discretization}.

\subsubsection{Motivation}

We put $\mathcal{D} = \{d: \mathcal{S} \to \mathbb{R}_{\geq 0}\}$ to be the set of distributions over the state space, and we consider functionals $\mathcal{R}(\cdot)$ that describe how desirable a given state distribution is. If we take $\mathcal{R}(d) = \mathbb{H}(d)$ on a discrete space we recover the formulation of \ref{ss:algo}. More generally, if we have some target distribution $d^*$, we can consider the reverse KL:
\begin{align}
    \mathcal{R}_{d^*}(\cdot) = D_{KL}(\cdot, d^*)
\end{align}

We consider the family of measurable surjective maps: $\mathcal{F}: \{\mathcal{S} \to \mathcal{Z}\}$, where $\mathcal{Z}$ is a discrete space of dimensions $k \leq dim(\mathcal{S})$. The inverse set map partitions $\mathcal{S}$, giving rise to the following relation:
\begin{align}
    d_1 \sim^f d_2 \iff f_*(d_1) = f_*(d_2)
\end{align}
where $f_*()$ is the push-forward measure on $\mathcal{Z}$. We let:
\begin{align}
    [\mu]_f = \{d \in \mathcal{D}: d \sim^f \mu\} = f^{-1}(Unif_z)
\end{align}

denote the subset of state distributions whose projection onto $\mathcal{Z}$ by $f$ result in the uniform distribution.

Then we put:
\begin{align}
    \mathcal{R}_f(\cdot) = D_{KL}(\cdot, [\mu]_f) = \mathbb{H}[f(\cdot)]
\end{align}

In words, we think about $f$ as a filter that represents various invariants on our state space $\mathcal{S}$. All state distribution that map uniformly on $\mathcal{Z}$ by $f$ are considered equally desirable, and we reward trajectories with high information content from the \emph{lens} of $f$.

\subsubsection{Learning the latent Representation} \label{ss:latent}

We let $\mathcal{Z}$ arise as a discrete bottleneck $\mathcal{S} \to^e \mathcal{Z} \to^d \mathcal{S}'$ from aiming to reconstruct state as best as possible, thereby inducing a representation $\mathcal{Z}$ that maximizes mutual information with $\mathcal{S}$. A smoothly parameterizes encoder (e.g. neural network) will ensure that similar states get mapped to similar codes. Hence, maximizing information content on $\mathcal{Z}$ will further encourages diverse trajectories on $\mathcal{S}$.

To discretize the latent space, we use \emph{locality sensitive hashing} (LSH) \cite{lsh}. LSH is helpful because state dimensions that are highly coupled (from an information-theoretic view), will be hashed to the same latent code by our auto-encoder whose projections to a lower dimension filter out small perturbations. On the other hand, a novel experiences will project to different latent subspace and thus will be prescribed a unique hash code. This approach decouples the ICE procedure from the state-space dimensionality, and can be directly applied to continuous spaces. 

We follow the discretization pipeline similar to \cite{count_based}, using a \emph{Simhash} \cite{simhash} scheme with a auto-encoder. Namely, for an encoder $e: \mathcal{S} \to \mathbb{R}^D$, a decoder $d: \mathbb{R}^D \to \mathcal{S}$, fixed stochastic matrix $A \in \mathbb{R}^{k \times D}$ with $A_{ij} \sim \mathcal{N}(0, 1)$, we retrieve the discrete latent code for state $s$ according to:
\begin{equation} \label{eq:ae_eq}
z(s) = sgn(A \sigma(\mathcal{U}_{(-a, a)} + e(s))) \in \{-1, 1\}^k
\end{equation}

where $\sigma(\cdot)$ denotes element-wise application of the sigmoid function, and $\mathcal{U}_{(-a, a)}$ is the injection of uniform noise to force the latent space to behave discretely \cite{binarized_nn}, \cite{relax_discrete}.

The encoder and decoder are parameterized by convolution neural networks, jointly trained with the reinforcement learning policy on a lagged updated schedule to ensure the latent codes are stable while retaining the ability to adjust based on exploratory progress of the agent. The auto-encoder optimizes a log likelihood term and an auxiliary loss that ensures unused latent bits take on an arbitrary binary value, preventing spurious latent code fluctuations \cite{count_based}:
\begin{equation} \label{eq:ae_loss}
    \mathcal{L}(\{s_i\}_{i=1}^N) = -\frac{1}{N}\sum_{i=1}^N[log p(s_i) - \newline
    \frac{\lambda}{K}\sum_{i=1}^Dg(s_i)]
\end{equation}
\begin{equation}
    g(s_i) := min\{(1-e(s_i), e(s_i)\}
\end{equation}

The algorithm for running ICE on a learned latent space is given in Alg.\ref{latent:algo}. We note that using learned hash codes involves minimal changes to the base ICE algorithm, requiring only that (1) states pass through the encoder and \emph{SimHash} prior to computing ICE reward, and (2) a buffer of approximately reconstructed state trajectories is stored to periodically updating the auto-encoder based on Eq.\ref{eq:ae_loss}.

\begin{algorithm}[tb]
   \caption{Information Content-based Intrinsic Reward in Learned Latent Space}
   \label{latent:algo}
\begin{algorithmic}
   \STATE {\bfseries Input:} Model \(M\), Encoder $e: \mathcal{S} \to \mathbb{R}^D$, Decoder $d: \mathbb{R}^D \to \mathcal{S}$, Fixed Gaussian matrix $A \in \mathbb{R}^{k \times D}$. Buffer $\mathcal{B}$
   \REPEAT
       \STATE {\bfseries Input:} state \(s_{0}\), {$H_{last} \gets 0$}, {$t \gets 0, \ done \gets False$}
       \STATE {\bfseries Input:} unique count dictionary \(q\)
       \STATE {$z_0 \gets Encode(s_0, e, A)$}, from Eq.\ref{eq:ae_eq}
       \STATE {$\hat{s_0} \gets Decode(s_0, z_0, d)$}
       \STATE {$\mathcal{B} \gets (s_0, z_0, \hat{s_0})$} add to buffer
       \STATE {$q \gets Update(q, z_{0})$}, from Eq.\ref{eq:unique_count}
       \STATE {$H_{last} \gets CalculateICE(q)$},  from Eq.\ref{eq:traj_entropy}
       \WHILE {not \(done\)}
            \STATE {$a_{t} \gets M(s_{t})$}
            \STATE {$s_{t+1},r^{extrinsic}_{t},done \gets env.step(a_{t})$}
            \STATE {$z_{t+1} \gets Encode(s_{t+1}, e, A)$}, from Eq.\ref{eq:ae_eq}
            \STATE {$\hat{s_{t+1}} \gets Decode(s_{t+1}, z_{t+1}, d)$}
            \STATE {$\mathcal{B} \gets (s_{t+1}, z_{t+1}, \hat{s_{t+1}})$} add to buffer
            \STATE {$q \gets Update(q, z_{t+1})$}
            \STATE {$H_{current} \gets CalculateICE(q)$}
            \STATE {$r^{intrinsic}_{t} \gets H_{current}-H_{last}$}
            \STATE {$r_{t} \gets r^{extrinsic}_{t} + \beta r^{intrinsic}_{t}$}
            \STATE {$H_{last} \gets H_{current}$}
            \STATE {$t \gets t+1$}
       \ENDWHILE
       \STATE {Update model \(M\)}
       \IF{iteration mod $ae_{update} = 0$}
        \STATE{update encoder $e$ and decoder $d$ with data from buffer $\mathcal{B}$ according to Eq. \ref{eq:ae_loss}}
       \ENDIF
   \UNTIL{Model \(M\) Converge}
\end{algorithmic}
\end{algorithm}

\section{Experiment}

We compare information content based intrinsic reward with several other methods from the literature described in Section \(2\). The selected environments include Grid-World, along with a group of sparse reward Atari games. All of the environments in this experiment are reduced to \(40 \times 40\) dimensionality and discretized using 255 gray scale range. We use a single convolution neural network with LSTM layers and seperate value and policy heads to parameterize the actor and critic. Details are given in \ref{appendex_net}.

\noindent The unique count matrix \(Q\) (Eq.\ref{eq:unique_count}) is preset to be \(Zeros(40,40,C)\) at the beginning of each trajectory, where \(C\) is the number of unique values for each pixel. The algorithm will dynamically update the matrix \(P\) and then calculate the trajectory information content after each step.

\subsection{Grid-World} \label{ss:gridworld}
\begin{figure}[ht]
    \centering
    \includegraphics[width=0.35\textwidth]{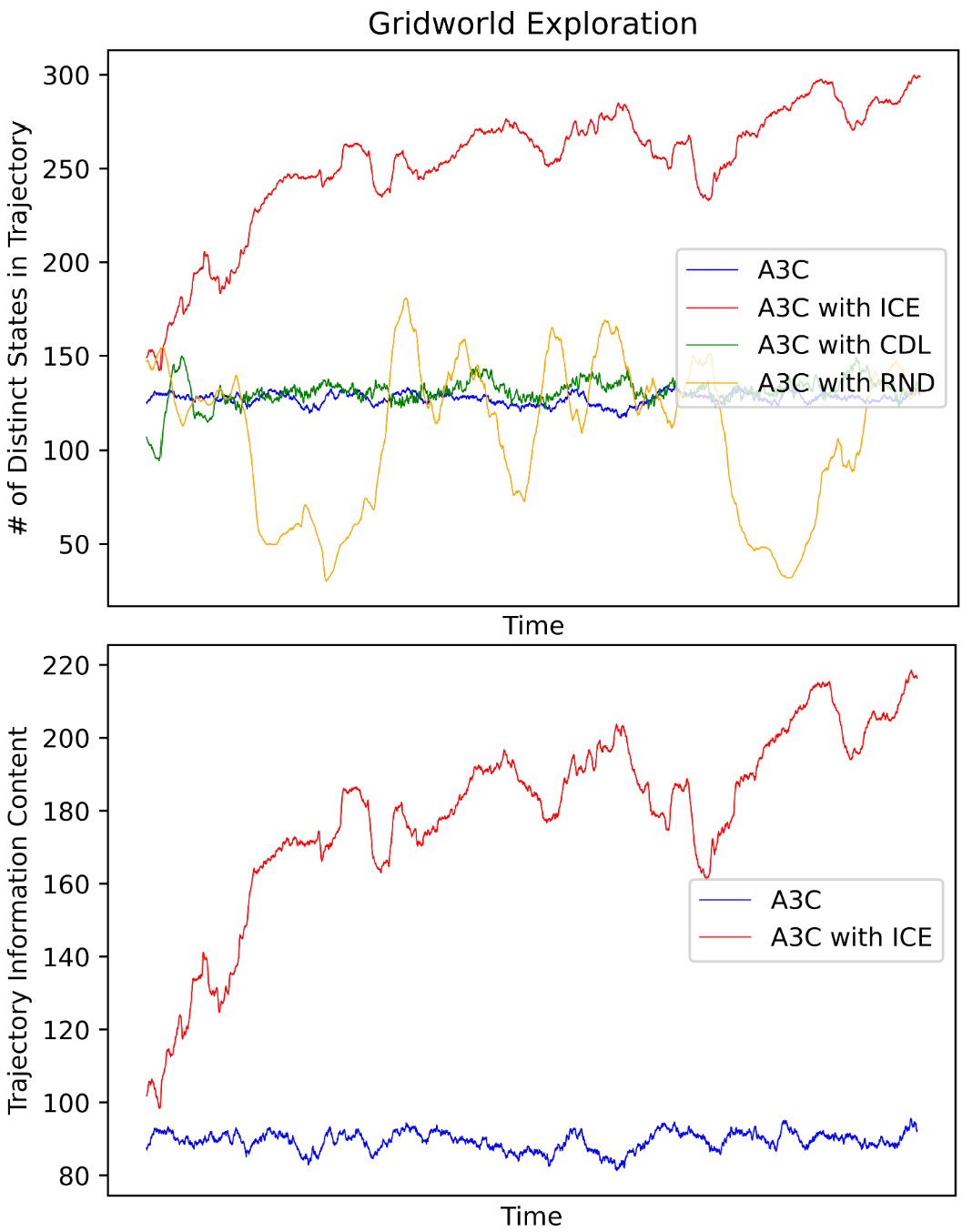}
    \caption{ICE motivates agent to reach a significantly higher number of distinct states which is correlated with an increase in information content. Measured in a no-reward grid-world environment.}
    \label{fig:gridworld}
\end{figure}

The state space for grid world is a \(40 \times 40\) grid (Fig.\ref{fig:gridworldwallmap} left). The agent starts at the (0,0) position and it can take action from \(\{up, down, left, right\}\). The grid visited by the agent is labeled as 1, otherwise 0, the extrinsic reward by the environment is always 0, and the environment terminates after 400 steps. \\ 
\noindent As shown in (Fig.\ref{fig:gridworld}), the agent is able to explore up to 300 distinct states in every trajectory with ICE enabled, which is significantly more than the case when ICE is disabled. The trajectory information content also increases as the number of distinct states increases. 

\subsection{Atari}
\begin{figure}[ht]
    \centering
    \includegraphics[width=0.48\textwidth]{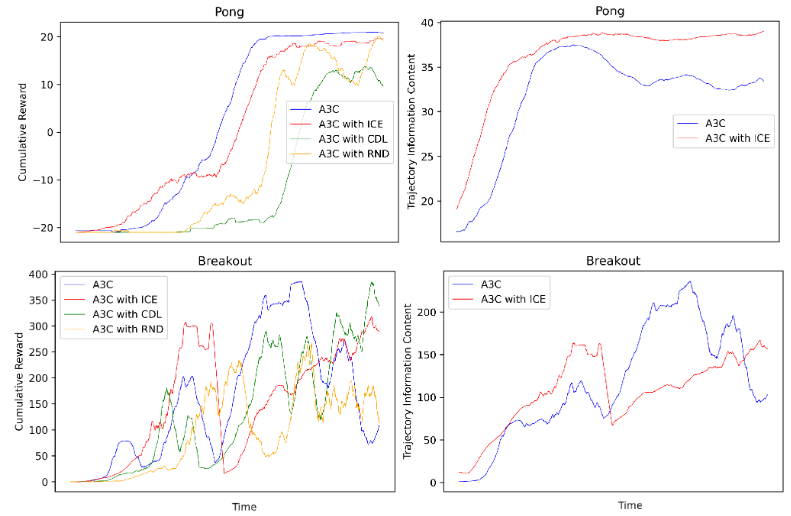}
    \caption{Performance on simple Atari Environments, cumulative reward (left), information content (right)}
    \label{fig:Atari}
\end{figure}

\begin{figure}[ht]
    \centering
    \includegraphics[width=0.48\textwidth]{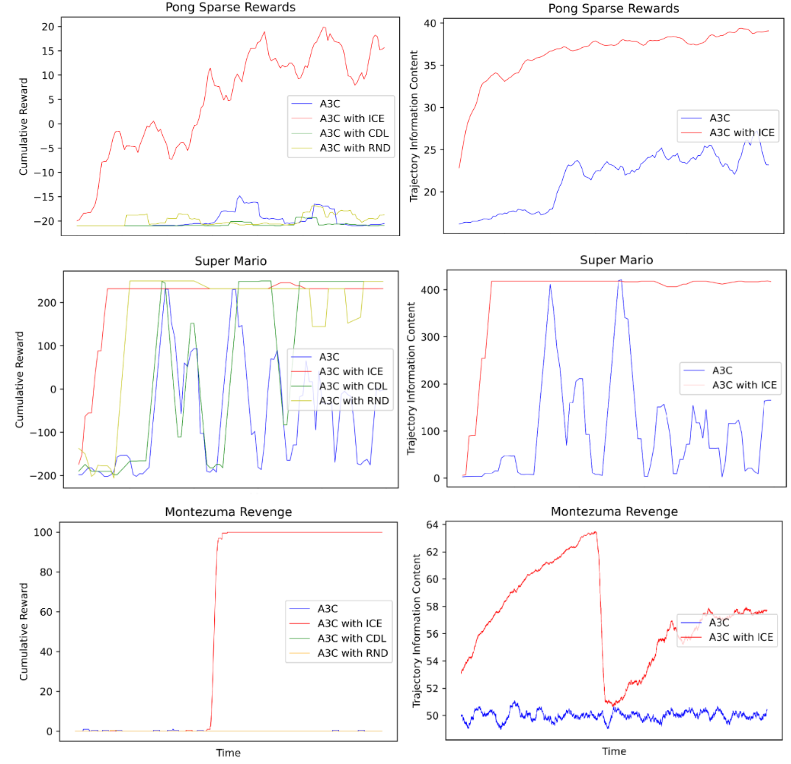}
    \caption{Performance on Sparse Reward Environments, cumulative reward (left), information content (right).}
    \label{fig:Atari-hard}
\end{figure}

We compare ICE with A3C, CDL, and RND on several environments from Atari. \\
Firstly we compare the methods of Pong and Breakout, Fig.\ref{fig:Atari}. The performance of ICE is slightly worse than A3C. This is because these two environments both have very rich reward systems. They do not necessarily require an exploration signal in order to solve. In these environments, the intrinsic rewards are noise rather than a signal to the model. Nonetheless, ICE will not degrade the performance by a lot as long as we keep it to a relatively small magnitude. As a result, ICE is able to solve both environments efficiently.
The second set of environments is Pong with Sparse Reward, Super Mario, and Montezuma, Fig.\ref{fig:Atari-hard}. These are challenging for RL algorithms due to their extremely sparse reward system. In Pong with Sparse Reward, all of the rewards were accumulated and provided to the model at the last step. In Super Mario, the agent gets no reward during learning and gets properly rewarded during evaluation (see \ref{ss:super mario} for more details). In Montezuma, the agent will only get a positive reward if the character reaches a key. These environments require the agent to precisely follow an action trajectory in order to solve the problem. In all cases, the ICE agent outperformed the base A3C agent. The learning for ICE is also very efficient as focuses on efficiently exploring the state space. ICE also displays interesting behaviors in the Montezuma Revenge environment. In the early stages, the agent focuses on exploring distinct states as it receives no rewards. As soon as it lands on a key, it quickly shifts its focus towards maximizing the extrinsic reward. Simultaneously, the intrinsic reward quickly degrades. Shortly after, the agent tries to regain the intrinsic reward while still being able to reach the key. As shown in the right column of Fig.\ref{fig:Atari},\ref{fig:Atari-hard}, ICE is very correlated with the environment's reward system, because Atari games such as Pong and Breakout have deterministic and observable state spaces, which gives a clear measure of information content using ICE. In general, ICE is a very robust exploration method because it is not very sensitive to hyper-parameters and is not prediction-based. 

\subsection{Grid-World with Wall} \label{ss:gridworldwall}
\begin{figure}[ht]
    \centering
    \includegraphics[width=0.48\textwidth]{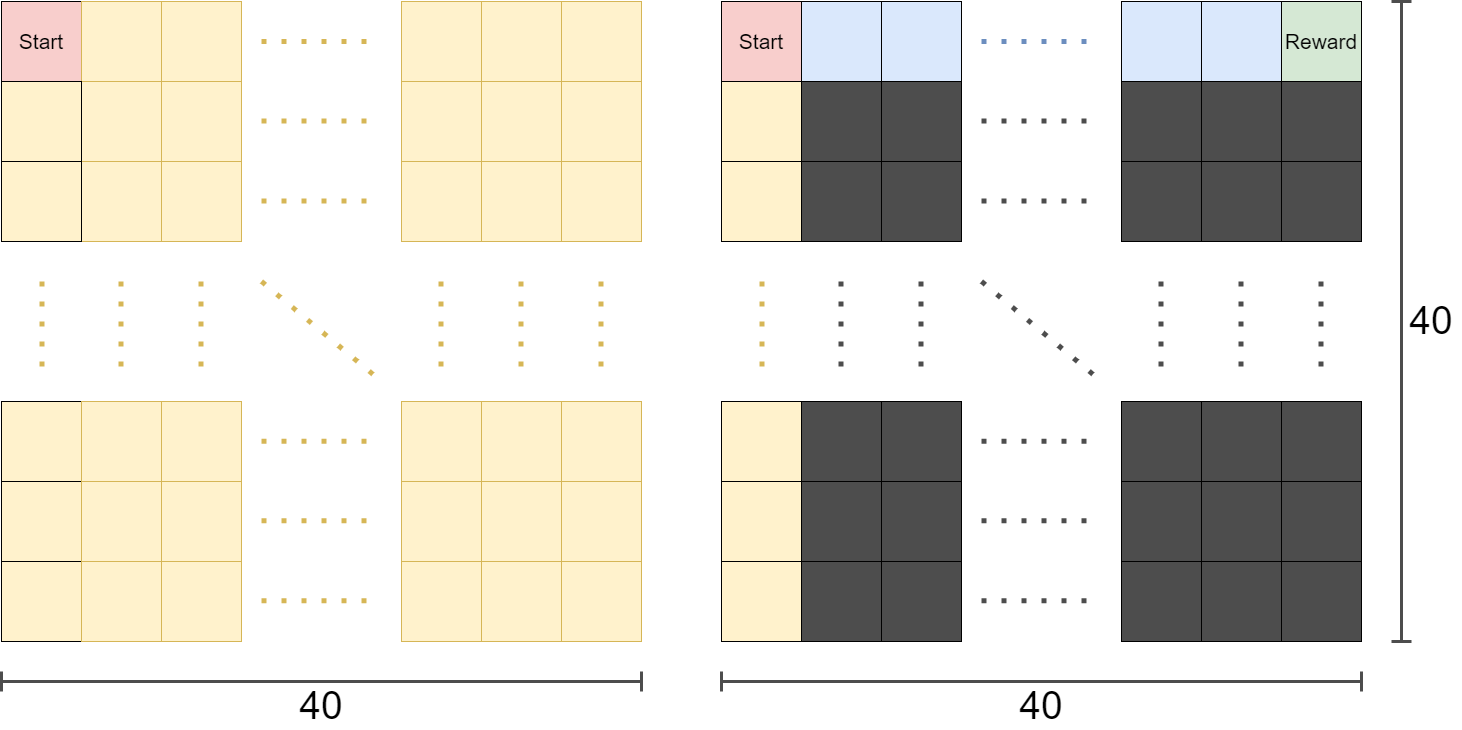}
    \caption{Setup of grid world (left) and grid world with wall (right). The agent will always start at the top-left corner. Yellow grids are observable, turn to 1 when visited. Blue grids are non-observable, always filled with 0. Black grids are walls that agents cannot step into. Green grid gives a positive reward to the agent.}
    \label{fig:gridworldwallmap}
\end{figure}

\begin{figure}[ht]
    \centering
    \includegraphics[width=0.48\textwidth]{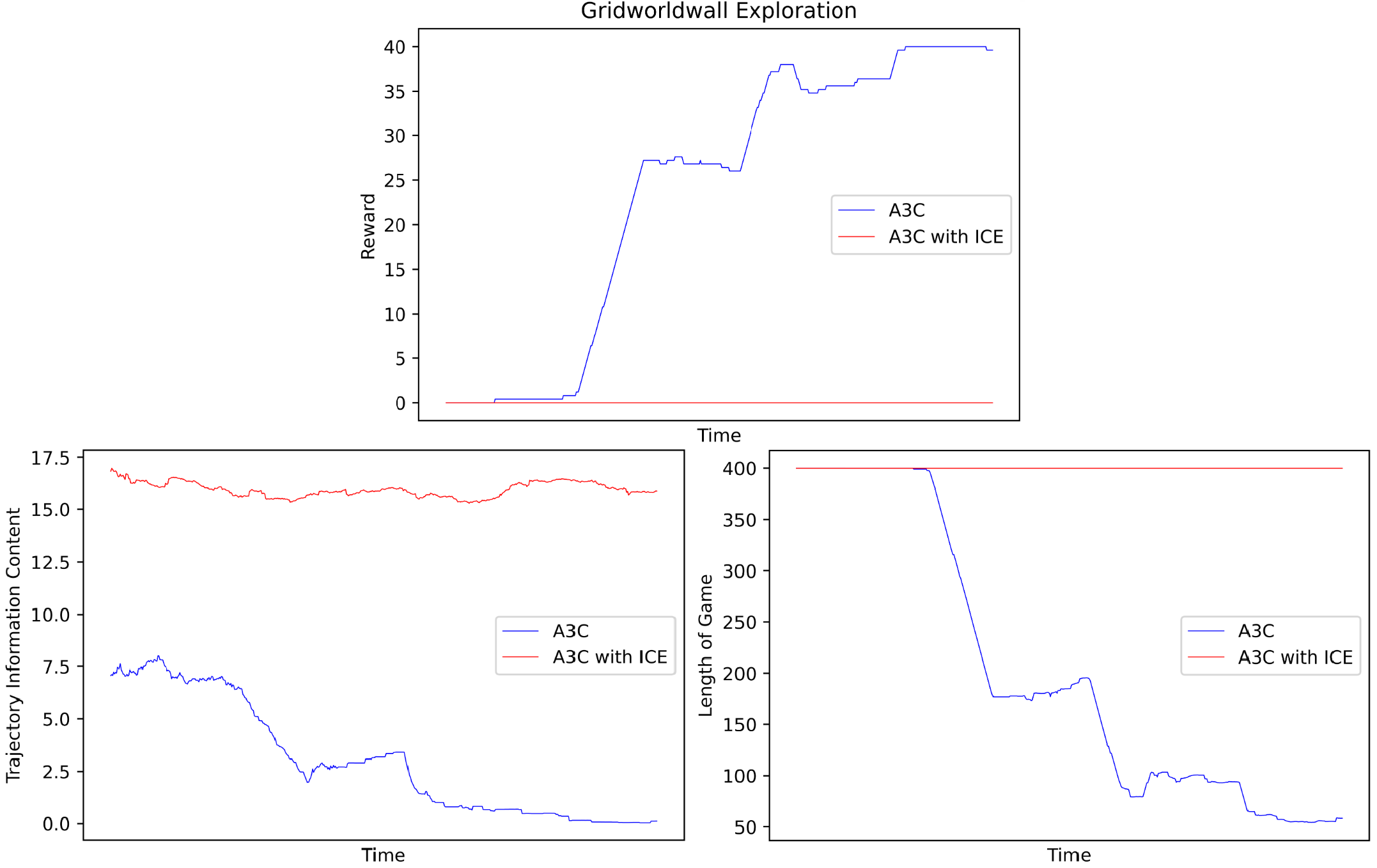}
    \caption{The agent with ICE is exploring efficiently in yellow grids but is not able to make it to the green grid. The agent without ICE finds the green grid after a few rounds of random search and quickly converges to this path.}
    \label{fig:gridworldwall}
\end{figure}

This is a special case of the Grid-World to illustrate the idea of \ref{ss:discussion}. The basic setup is the same as described in \ref{ss:gridworld}, with differences shown in Fig.\ref{fig:gridworldwallmap}. The block of the wall marked in black represents states the agent cannot visit. On top of this, the grids marked in blue are unobservable, as they will always be filled with 0, and the grids marked in yellow are standard (turns to 1 if visited, else 0). The agent will get a positive reward and terminate the game if it has reached the green grid. The agent with ICE is efficiently exploring yellow grids, but it is having trouble exploring blue grids and cannot land on the green grid. This is because the agent quickly converged to take the path in yellow grids, as it can generate higher information content compared to a path in blue grids. On the other hand, the agent without ICE is able to land at the green grid after a while of random search (Fig.\ref{fig:gridworldwall}). This is a limitation of ICE that can be overcome by simply including action exploration.  \\ 

\subsection{ICE in a Learned Latent Space} \label{test:discretization}

We evaluate how well ICE performs when computed on a learned latent space, following the procedure of \ref{ss:latent}, on several environments including Pong, Breakout, Montezuma's revenge and Starpilot. The encoder consists of 3 convolution blocks followed by a linearity outputting a $128$-dimensional dense representation which is hashed into $k=16$-dimensional latent codes. The decoder consists of 3 deconvolution blocks. We use $\lambda=0.5$ for the auxiliary auto-encoder loss, and update the auto-encoder every 3 updates of the base policy. This architecture is inspired by the representation learning stack from \cite{count_based}, which aim to achieve high reconstruction accuracy to help stabilize the learned latent space, and thereby make our ICE reward more consistent.

The extrinsic reward learning curves are shown in \ref{ss:latent}. We observe that working in a compressed representation can be beneficial, even for state spaces that are already discrete (e.g. images). In Pong and Starpilot, ICE in latent space improves the speed of learning, suggesting that the bottlecked representation helps regularize against sporadic noise in our state trajectories.

In Montezuma's revenge, optimizing for ICE in latent space exhibits asymptotically better extrinsic rewards, suggesting that maximizing information content on the compressed representation is better able to capture semantically meaningful subspaces worth exploring. 

We believe that these initial results are promising, and we are motivated to further experiment with ICE rewards on various representation spaces.

\begin{figure}[ht]
    \centering
    \includegraphics[width=0.48\textwidth]{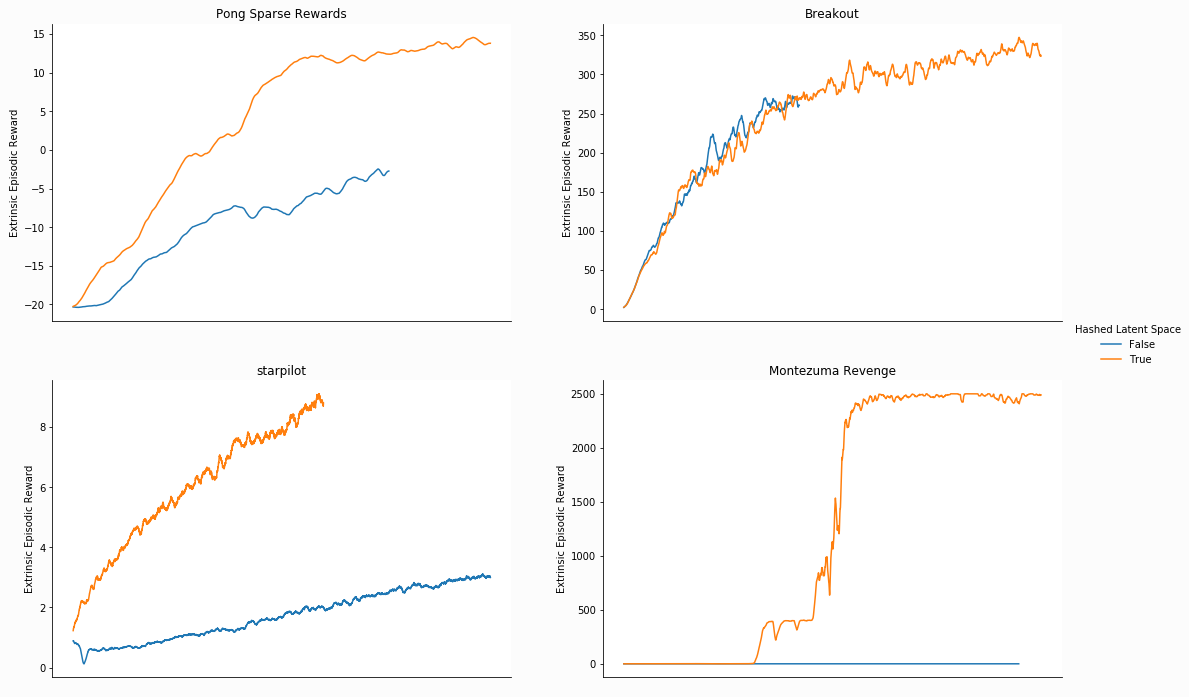}
    \caption{Comparing performance of ICE reward in learned latent space (orange) vs. ICE reward directly on uniformly discretized pixels (blue). Optimizing the ICE reward on the learned latent space improves sample-efficiency in Pong, Starpilot and Montezuma.}
\end{figure} 

\section{Conclusion}
We have introduced a new intrinsic reward based on the information content of trajectories measured using Shannon’s Entropy. Our method is amenable to vectorized implementation and has shown robustness in various observable MDP's. ICE motivates the agent to explore diversified trajectories, thereby increasing its sample-efficiency in sparse environments where the probability of receiving a reward is directly proportional to the number of distinct visited states. We discussed a discretization scheme \ref{ss:discretization} based on locality sensitive hashing that maximizing information content on a compressed latent space, generalizing ICE reward to continuous state spaces. We also examined the limitations of motivating agents to traverse trajectories with a large number of distinct states in \ref{ss:gridworldwall}.

\section{Future Work}

The idea of this work is loosely related to depth-first search (DFS). By analogy, action state exploration is a cyclic breadth-first search (BFS). We believe that there is opportunity to improve ICE by dynamically shifting between DFS and BFS behaviour as the agent interacts with the environment. We also posit that changing the formulation to incorporate information \emph{across} multiple episodes can yields better results. For example, we can maximize information content of the current trajectory while minimizing the mutual information with previous experience; we leave this to future work. Finally, we've observed promising results using ICE on a learned latent representation. In future work, we wish to further validate this process by evaluating it using different representations such as the \emph{successor representation} \cite{ssr}, and rigorously testing sensitivity to hash-dimension on continuous spaces.

\section*{Software and Data}
We include an A3C and a PPO implementation of ICE in the Supplementary Material.

\bibliography{ice}
\bibliographystyle{icml2023}

\newpage
\appendix
\onecolumn
\section{Appendix}

\subsection{Hyperparameters}

\begin{table}[t]
\caption{Hyperparameters used in our experiments.}
\label{sample-table:hyperparameter}
\vskip 0.15in
\begin{center}
\begin{small}
\begin{sc}
\begin{tabular}{lccccr}
\toprule
& gridworld & gridworldwall & Pong & Montezuma & Mario \\
\midrule
    lr  & \(10^{-4}\) & \(10^{-4}\) & \(10^{-4}\) & \(10^{-4}\) & \(10^{-4}\) \\
    gamma & 0.99 & 0.99 & 0.99 & 0.99 & 0.99\\
    \(\alpha_{value}\) & 0.5 & 0.5 & 0.5 & 0.5 & 0.5\\
    \(\alpha_{policy}\) & 1.0 & 1.0 & 1.0 & 1.0 & 0.5 \\
    \(\alpha_{entropy}\) & 0.01 & 0.01 & 0.01 & 0.01 & 0.01 \\
    \(\beta\) & 0.5 & 0.5 & 0.5 & 1.0 & 0.1 \\
\bottomrule
\end{tabular}
\end{sc}
\end{small}
\end{center}
\vskip -0.1in
\end{table}

\subsection{1-D Random Walk Analysis} \label{ss:a2}
\begin{figure}[ht]
    \centering
    \includegraphics[width=0.5\textwidth]{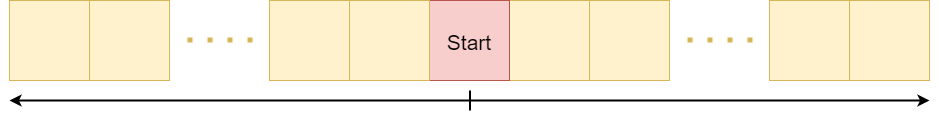}
    \caption{Setup of a 1-D random walk grid. The agent starts at the 0 position and takes action from {left, right} randomly.}
    \label{fig:randomgrid}
\end{figure}

We want to show by calculation that it is very unlikely an agent can end at equal or larger than \(K\) distances away from the origin after taking \(N\) random steps in a 1-dimension grid. As shown in Fig.\ref{fig:randomgrid}, the agent starts at the 0 position and takes action from {left, right} randomly. Define the probability of an agent end at equal or larger than \(K\) distances away from the start position after taking \(N\) random steps to be \(P_{k}^{N}\). \\
In the case where \(N \geq K\), it follows from the binomial distribution,
\begin{equation} \label{eq:randomwalk}
P_{k}^{N} = \sum_{i=\left\lceil \frac{K+N}{2} \right\rceil}^{N} \binom{N}{i} \left(\frac{1}{2} \right)^{N}
\end{equation}
For large \(N\) and \(K \ll N\), we can apply Stirling’s Formula \cite{dutka1991early} to simplify the calculation.

\subsection{More examples} \label{ss:example}
\noindent We will illustrate \(r^{intrinsic}_{t}\) by an example. Consider a simple grid world environment with dimension \(1 \times 4\). Every grid begins with value \(0\), and changes to 1 on visitation. The agent starts at the leftmost grid. The agent has \(2\) possible actions (left, right). The environment itself provides no reward, \(r^{extrinsic}_{t} = 0, \forall \ t\). The game terminates after \(3\) steps. Let's consider an example trajectory, where the actions are \(left, right, right\).
\begin{equation}
s_{0}, s_{1}, s_{2}, s_{3} = [1,0,0,0], [1,0,0,0], [1,1,0,0], [1,1,1,0]
\end{equation}
\noindent Then the trajectory for each state variable up to \(t=4\) is
\begin{equation}
X^{d=0}_{t=3}, X^{d=1}_{t=3}, X^{d=2}_{t=3}, X^{d=3}_{t=3} = [1,1,1,1], [0,0,1,1], [0,0,0,1], [0,0,0,0]
\end{equation}
\noindent The count unique and probability occurrence vector for each state variable up to \(t=3\) is
\begin{equation}
\begin{split}
& q^{d=0}_{t=3}, q^{d=1}_{t=3}, q^{d=2}_{t=3}, q^{d=3}_{t=3} = [0,4], [2,2], [3,1], [4,0] \\
& p^{d=0}_{t=3}, q^{d=1}_{t=3}, q^{d=2}_{t=3}, q^{d=3}_{t=3} = [0/4,4/4], [2/4,2/4], [3/4,1/4], [4/4,0/4]
\end{split}
\end{equation}
\noindent The total information content of this trajectory can be calculated as
\begin{align*}
\begin{split}
& H^{d=0}_{t=3}, H^{d=1}_{t=3}, H^{d=2}_{t=3}, H^{d=3}_{t=3} = 0, 1, 0.81, 0  \ bits \\
& H_{t=3} = \sum H^{d}_{t} = 1.81 \ bits \\
\end{split}
\end{align*}
\noindent Other \(H_{t}\) where \(t \in [0,1,2]\) can be calculated through the same method, where \(H_{t=0}=0, H_{t=1}=0, H_{t=2}=0.92\). Lastly, the intrinsic reward is
\begin{equation}
r^{intrinsic}_{t=3} = H_{t=3}-H_{t=2} = 0.89 \ bits
\end{equation}
Note that the agent can increase its information content by visiting more distinct grids within the trajectory, and the maximum is achieved when all grids have been visited exactly once within \(N\) steps. The grid world environment is a very representative environment for many sparse reward environments. They are similar in a way that their exploration objective can all be described as increasing the probability of reaching a non-zero reward state by exploring more distinct states.

\subsection{Super Mario Reward System} \label{ss:super mario}
\noindent The extrinsic reward for the Super Mario environment is always 0 during learning, and the extrinsic reward during evaluation is as follows \cite{gym-super-mario-bros},
\begin{equation}
\begin{split}
& velocity_{t} = position_{t} - position_{t-1} \\
& clock_{t} = time_{t-1} - time_{t} \\
& death_{t} = 0 \ if \ alive \ else \ -15 \\
& r^{extrinsic}_{t} = velocity_{t} + clock_{t} + death_{t}
\end{split}
\end{equation}

\subsection{Neural Network Architecture} \label{appendex_net}

\begin{figure}[ht]
    \centering
    \includegraphics[width=0.45\textwidth]{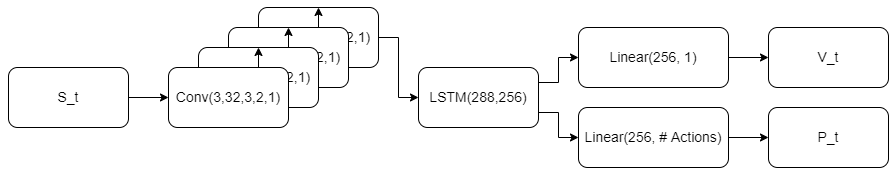}
    \caption{Neural Network Structure}
    \label{fig:NN Network}
\end{figure}

The neural network is designed to be a combination of 4 convolution, 1 LSTM, and 2 linear layers (Fig.\ref{fig:NN Network}). The LSTM layer is important here to give the model a sense of the states it has traveled so far in a trajectory.

\subsection{Locality Sensitive Hashing} \label{appendix_simhash}

We consider \emph{locality sensitive hashing}. A family of hash functions $\mathcal{H} = \{h: \mathcal{S} \to \mathbb{Z}\}$ is \emph{locality sensitive for a similarity function s}: $\mathcal{S} \times \mathcal{S} \to \mathbb{R}$ if a hash function $h \sim \mathcal{H}$  randomly sampled from the family satisfies:
\begin{align}
    d(s_1, s_2) &\leq R \implies p(h(s_1) = h(s_2) \geq P_1) \\
    d(s_1, s_2) &\geq C \cdot R \implies p(h(s_1) = h(s_2)) \leq P_2
\end{align}
for some constants $C>1, R>0$ and for probabilities $P_1, P_2 \geq 0$. In other words, points close together are hashed to the same value with high probability, and dissimilar points are unlikely to collide in hash. Implicit in this assumption is that $(\mathcal{S}, d)$ form a valid metric space.

Locality sensitive hashing enables us to map a continuous, high dimensional state space into a compressed latent space such that similar points in state space map to similar latent codes.
\end{document}